\pdfoutput=1
\documentclass[letterpaper, 10 pt, conference]{ieeeconf}  

\usepackage{blindtext, graphicx}
\usepackage{gensymb}
\usepackage{xcolor}

\usepackage{tikz,lipsum}
\usepackage[labelformat=simple]{subcaption}

\DeclareCaptionLabelSeparator{periodspace}{.\quad}
\usepackage{listings}

\usepackage{algorithm}
\usepackage{algorithmic}

\usepackage{ amssymb }
\usepackage{amsmath}
\usepackage{commath}

\usepackage{framed} 

\usepackage{float}
\usepackage{url}

\usepackage{array}
\newcolumntype{P}[1]{>{\centering\arraybackslash}p{#1}}
\newcolumntype{M}[1]{>{\centering\arraybackslash}m{#1}}

\usepackage{url}

\usepackage{breakurl}
\usepackage[breaklinks]{hyperref}

\IEEEoverridecommandlockouts                              

\title{\LARGE \bf
Lio - A Personal Robot Assistant for Human-Robot Interaction and Care Applications
}

\author{Justinas Mi\v{s}eikis, Pietro Caroni, Patricia Duchamp, Alina Gasser, Rastislav Marko, Nelija Mi\v{s}eikien\.e,\\ Frederik Zwilling, Charles de Castelbajac, Lucas Eicher, Michael Fr\"{u}h, Hansruedi Fr\"{u}h
\thanks{Authors are with F\&P Robotics AG, Z\"{u}rich, Switzerland, {{\tt\small info@fp-robotics.com}}, {{\tt\small www.fp-robotics.com}}}
}

\begin{document}

\maketitle
\thispagestyle{empty}
\pagestyle{empty}

\begin{abstract}

Lio is a mobile robot platform with a multi-functional arm explicitly designed for human-robot interaction and personal care assistant tasks. The robot has already been deployed in several health care facilities, where it is functioning autonomously, assisting staff and patients on an everyday basis. Lio is intrinsically safe by having full coverage in soft artificial-leather material as well as having collision detection, limited speed and forces. Furthermore, the robot has a compliant motion controller. A combination of visual, audio, laser, ultrasound and mechanical sensors are used for safe navigation and environment understanding. The ROS-enabled setup allows researchers to access raw sensor data as well as have direct control of the robot. The friendly appearance of Lio has resulted in the robot being well accepted by health care staff and patients. Fully autonomous operation is made possible by a flexible decision engine, autonomous navigation and automatic recharging. Combined with time-scheduled task triggers, this allows Lio to operate throughout the day, with a battery life of up to 8 hours and recharging during idle times. A combination of powerful on-board computing units provides enough processing power to deploy artificial intelligence and deep learning-based solutions on-board the robot without the need to send any sensitive data to cloud services, guaranteeing compliance with privacy requirements.  During the COVID-19 pandemic, Lio was rapidly adjusted to perform additional functionality like disinfection and remote elevated body temperature detection. It complies with ISO13482 - Safety requirements for personal care robots, meaning it can be directly tested and deployed in care facilities.

\end{abstract}

\section{INTRODUCTION}
\label{sec:introduction}

Recently robots have been gaining popularity outside the factory floors and entering unstructured environments such as homes, shops and hospitals. They range from small devices designed for \textit{internet-of-things (IoT)} applications to larger physical robots which are capable of autonomously navigating in indoor and outdoor environments, sharing the workspace with people and even interacting with them. 

Given the issue of ageing population and shortage of medical and nursing staff in many countries, this naturally leads to attempts to use robotics and automation addressing this problem~\cite{flandorfer2012population}~\cite{oliver2014making}. 
For example, in Switzerland, the number of people over 80 years of age is expected to double from 2015 to 2040~\cite{suisse2015zukunft}. It will result in nearly triple nursing costs for the Swiss healthcare sector.

Furthermore, healthcare employees are experiencing severe working conditions due to stress, underpayment and overtime. For example, Between $8$\% and $38$\% of health workers suffer physical violence at some point in their careers~\cite{WHOViole45:online}. A possible staff shortage of 500'000 healthcare employees is estimated in Europe by the year of 2030~\cite{Rothgang2014Themenreport}.

Care robotics is not an entirely new field. There has been significant development in this direction. One of the most known robots is \textit{Pepper} by \textit{SoftBank Robotics}, which was created for interaction and entertainment tasks. It is capable of voice interactions with humans, face and mood recognition. In the healthcare sector \textit{Pepper} is used for interaction with dementia patients~\cite{ikeuchi2018utilizing}.

Another example is the robot \textit{RIBA} by \textit{RIKEN}. It is designed to carry around patients. The robot is capable of localising a voice source and lifting patients weighing up to $80$ kg using remote control~\cite{goeldner2015emergence}. 

The \textit{Sanbot Elf} robot by \textit{QIHAN} has numerous sensors allowing it to monitor the health condition of patients and residents. A recent study with \textit{Sanbot Elf} even tested its capability to detect fallen elderly residents~\cite{bauer2018camera}. However, the robot is not capable of handling and manipulating objects. Another robot is \textit{WALKER} by \textit{UBTECH}. It can manipulate objects, climb stairs and do yoga~\cite{UBTECHSh74:online}.

The \textit{COVID-19} crisis has given an increase in robots applicable to the health care sector. Most of them are not able to manipulate objects, they are optimised for autonomous driving and delivery of objects. Robots like \textit{Peanut} from \textit{Keenon Robotics Co.}, the \textit{SAITE Robot} or the \textit{NOAH Robot} have been deployed to assist with it.

The \textit{Care-o-bot 4} by \textit{Unity Robotics} and \textit{Fraunhofer IPA} is used in four shopping centres. It can recognise faces, provide daily news and handle objects~\cite{kittmann2015let}. Despite the initial focus on the healthcare market, the existing applications are limited to pilot projects only.

One more healthcare oriented robot is \textit{Moxi} by \textit{Diligent Robotics}. Compared to previously described robots, the focus of \textit{Moxi} is not interacting with people, but rather lies in the logistic of hospitals. The robot can deliver medical samples, carry laundry, bring supplies and greet patients~\cite{ackerman2018moxi}. It uses a mobile platform, arm, gripper and an integrated lift to adjust its height. \textit{Moxi} is mostly capable of completing the tasks in an end-to-end manner. Door opening and closing procedures are not in the range of its capabilities, it relies on automatic door systems or places the packages outside the room.

\begin{figure}[ht]
    \vspace{0.2cm}
    \centering
    \includegraphics[width=0.49\textwidth]{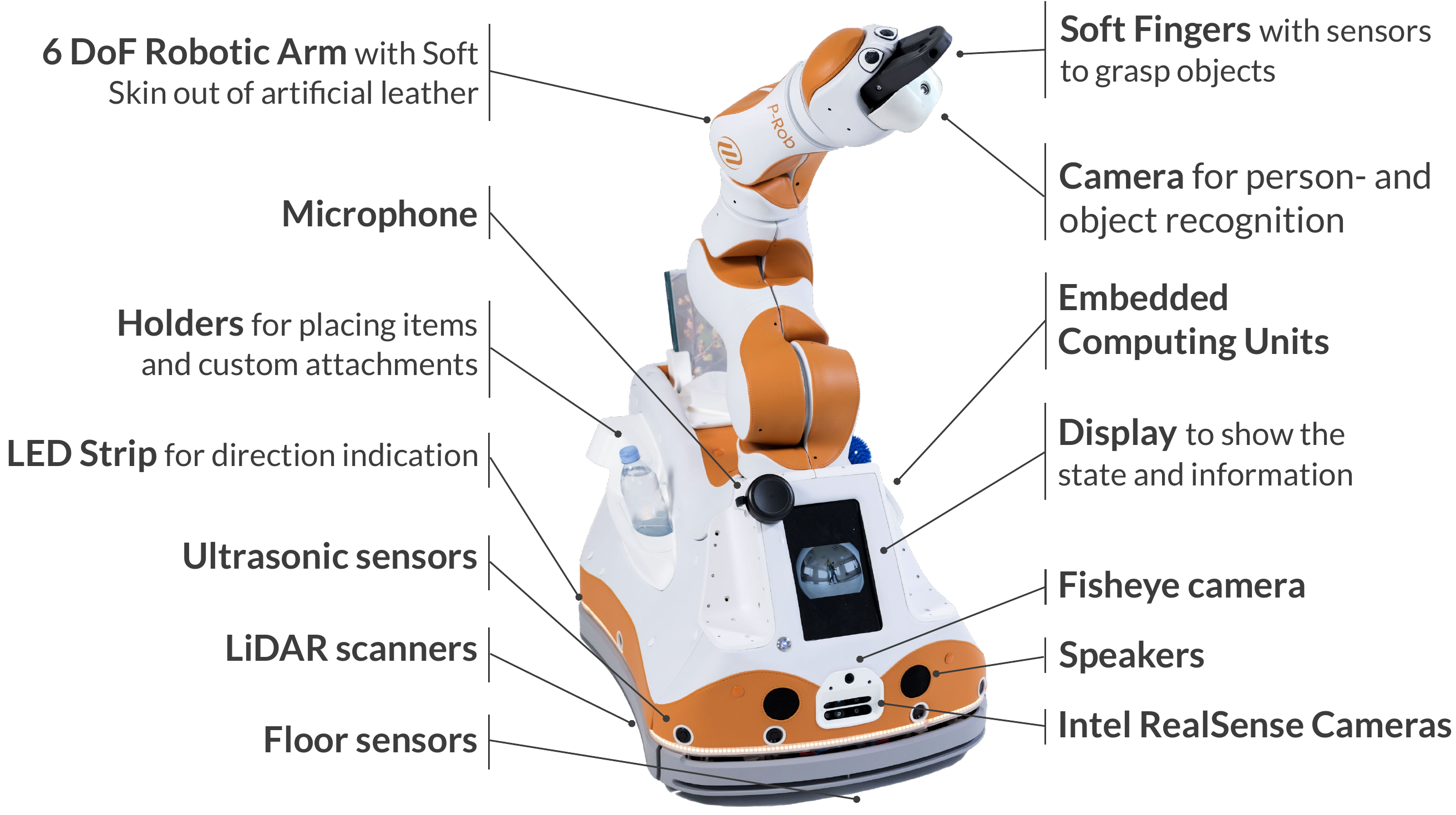}
    \caption{Overview of Lio: sensor and hardware setup.}
    \label{fig:lio_explained}
    \vspace{-0.6cm}
\end{figure}

Furthermore, the robots of PAL Robotics, primarily the \textit{REEM-C} and \textit{TIAGo} robots, could be used in the healthcare sector. They are developed with research and industry in mind but no specific application. Open platforms allow users to freely explore possibilities, such as gesture recognition~\cite{pfeiffer2015gesture}.

In the market of social robots, there are more platforms focused on interaction without manipulation capabilities such as \textit{PARO}, \textit{KIKI}, \textit{AIBO}, \textit{Buddy}, \textit{Kuri}, \textit{Mykie} and \textit{Cozmo}. While such robots often achieve their goal of creating positive emotions, or even therapeutic progress, they are incapable of assisting or solving tasks.

In this paper, a personal robot platform - Lio, shown in Fig.~\ref{fig:lio_explained}, is presented. The robot is specifically designed for autonomous operation in health care facilities and home care by taking into account the limitations of other robot platforms and improving upon it. Lio is intrinsically safe for human-robot interaction. It is the next iteration robot platform built upon the experience of creating a \textit{P-Care} robot, which was a well-accepted personal robot developed for Chinese markets~\cite{fruh2018erfahrungen}.

First of all, the system is described from the hardware point of view, explaining the functionality and capabilities of the robot. Then software, interfaces and existing algorithms are presented followed by the challenges of operating a robot in hospital environments. Eventually, existing use cases and deployments of Lio are presented incorporating evaluation and current limitations of the system. The paper is concluded with an overview of Lio in the context of personal care and research applications followed by current developments and future work.

\section{SYSTEM DESCRIPTION}
\label{sec:system_description}

Lio, a personal assistant robot for care applications can complete a wide variety of complex tasks. It can handle and manipulate objects for applications in health care institutions and home environments. Furthermore, it operates autonomously in an existing environment without requiring significant adaptations for its deployment. The design of Lio evolved as an iterative process. During the deployments, observations and interactions with staff and patients were taken into account for further development.

A set of heterogeneous sensors are embedded in the robot. It provides Lio with the capabilities of navigating, understanding and interacting with the environment and people. A combination of in-house developed robot control and programming software \textit{myP} together with algorithms based on the \textit{Robot Operating System (ROS)} introduces a node-based system functionality. It enables the communication between all the modules and an easy overview, control and interchangeability of all the software modules. For custom development, full access to the sensor data and control of the robot is provided.

\subsection{Robot Hardware Overview}
\label{subsec:sd_overview}

Lio consists of a robot arm placed on top of a mobile platform. The robot was designed by keeping in mind the workspace of the robot arm. It enables Lio to combine arm and mobile platform movements to grasp objects from the ground, reach table-tops and be at a comfortable height for interaction with people in wheelchairs. It was a necessary measure to ensure that the robot is not too tall in order not to be intimidating to people who are sitting.

Padded artificial-leather covers ensure that Lio will not cause injuries in the event of a collision but also provide significantly more cosy and friendly appearance compared to hard-shell robots. Majority of the body has a soft-feel to it, which helps with the acceptance of the robot.

In any health care environment, data privacy is at the utmost importance. Lio was specifically designed to have enough processing power to be able to run complex algorithms, including deep learning-based ones, entirely locally without the need for cloud computing or transferring data outside of the robot. Lio has four embedded computing units: \textit{Intel NUC}, \textit{Nvidia Jetson AGX Xavier}, \textit{Raspberry Pi} and an \textit{embedded PC with Atom processor}.

Each of the computing units is responsible for running separate modules with constant communication using an internal secure Ethernet network. Communication is based on \textit{ROS} topics, meaning that additional modules could be integrated into the system by using standard \textit{ROS} communication protocols and make use of sensor data.

\subsection{Mobile Platform and Navigation}
\label{subsec:sd_mobile_platform}

\begin{figure}
    \vspace{0.2cm}
    \centering
    \includegraphics[width=0.40\textwidth]{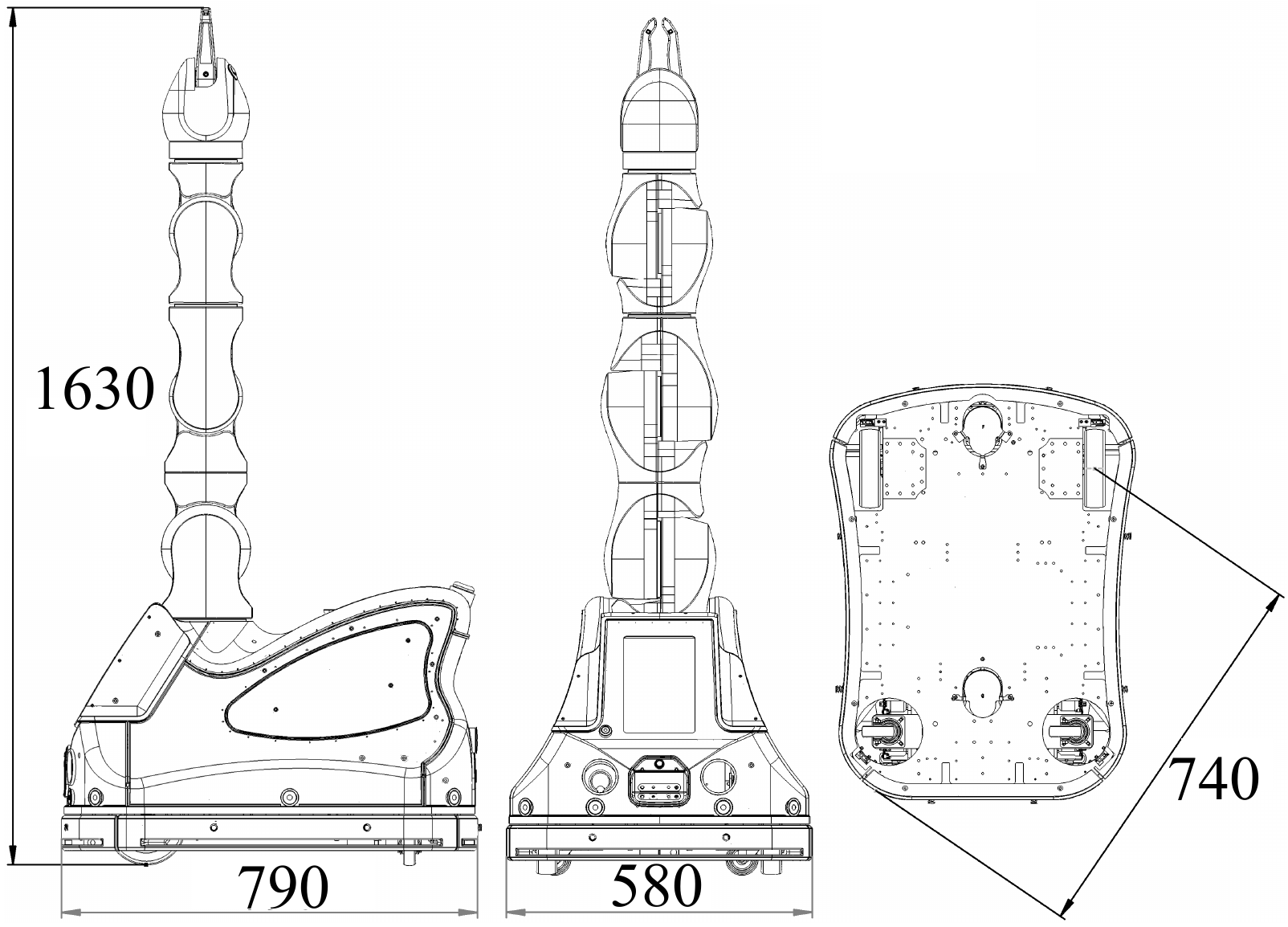}
    \caption{Technical drawing of Lio.}
    \label{fig:lio_tech_drawing}
    \vspace{-0.6cm}
\end{figure}

The mobile platform allows Lio to safely navigate in the environment and plan movement trajectories even in cluttered places. The mobile platform is $790$x$580$ mm in size, as shown in Fig.~\ref{fig:lio_tech_drawing}. Therefore, it is suitable to operate in any environment prepared for wheelchairs. The mobile platform uses a differential drive system with two caster wheels, allowing it to turn around in small spaces.

For safe navigation, different sensors are embedded in the mobile platform. Two LiDARs are placed behind the bumpers in the front and the back of the mobile platform for coverage around the robot. Their measurements are merged into a 360\degree~scan. Additionally, distance sensors are distributed around the circumference of the mobile platform for redundant proximity detection of the obstacles. In the scenario of Lio bumping into an obstacle, four mechanical bumpers ensure that the collision is detected and the robot will stop before any damage is caused. Four downward-facing infrared floor sensors were added to detect stairs or any edges. The robot moves at speeds up to $1$ m/s, ensuring short stopping distance when needed. To provide visual information, two \textit{Intel RealSense D435} Depth Cameras are embedded in the front. One camera is facing downwards to detect the floor plane, while the other one faces up for higher obstacle detection in front of the robot. Fields of view (FoV) of both cameras overlap fully and cover the frontal view. In addition to that, a wide-angle fisheye camera overviews the full frontal view with over 170\degree~FoV.

The map of the environment is created from the data of the two LiDARs. Mapping uses \textit{gmapping} from OpenSLAM package and a probabilistic localisation system is based on the adaptive Monte Carlo method~\cite{stachniss2007openslam}~\cite{fox1999monte}. Both of these methods were specifically adapted for Lio. On the map of the facility, there is a possibility to add a layer of virtual objects to prevent the robot from going to these areas. Lio can locate the charging station and navigate to it autonomously when running low on battery.

Custom positions can be saved on the map such as specific waypoints or general areas like kitchen, patient rooms and nursing stations can be added. All the sensor information is used to ensure safe navigation through the facilities for Lio. Furthermore, the LED strip placed around the mobile platform indicates the current driving direction and state of the robot allowing people to understand the intentions of Lio.

For interaction purposes, a non-touch display is embedded in the front of the mobile platform. Use case studies have shown that voice and touch interactions with the robot-arm are preferred over the touch screen interface~\cite{fischinger2016hobbit}. Mounting touch-screen devices low on the platform could potentially be a hazard due to bending down and risk of falling.

The display shows the status of the robot, displays text during the voice interaction and custom-designed visualisations that might be beneficial during the operation of the robot. Visualisation on the screen is fully customisable through the scripts. Additionally, robot camera feeds can also be visualised on the screen. Embedded loudspeakers, together with a multi-directional microphone, enable Lio to interact using voice and sound. Lio can understand voice commands as well as generate speech from text or play music.

For safety purposes, an emergency button is located on the back of the robot platform and is easily reachable in case the robot has to be suddenly stopped. When pressed, the robot and the mobile platform are released, making it easy to move by hand. Upon the release of the emergency button, Lio returns to the normal operation mode.

\subsection{Robot Arm}
\label{subsec:sd_robot_arm}

The robot arm placed on top of the mobile platform is a \textit{P-Rob 3}. It is a six-degree-of-freedom (DoF) collaborative robot arm designed for human-robot interaction tasks with the maximum payload of $3$-$5$ kg. \textit{Optical pseudo-absolute} encoders simplify the calibration task upon startup. The calibration can be executed from any position, takes under $3$ seconds, and each joint is moved by a maximum of $5$\degree.

A trajectory planner is offered by \textit{myP}, which differentiates between joint space and tool space paths. An extensive kinematics module handles singularities and all possible robot configurations. Both analytical and numerical calculations with various positional and directional constraints can be executed. High-frequency calculations are made possible by having the kinematics module as a standalone \textit{C++} library, which can also be imported into external projects.

For increased safety, the \textit{Motion Control Module} (\textit{MCM}) includes a compliant position control mode. A feed-forward \textit{proportional-derivative (PD)} controller with adjustable gain and stiffness creates a soft motion behaviour. The mode is particularly useful when interacting with people as well as upon contact with rigid real-world objects such as doors. Compliant control is used for human-robot interaction tasks as an input option. For example, users can push the head of the robot to initiate actions or provide a positive answer, pushing the head side-to-side indicates a negative answer. Safety is ensured by limited joint accelerations and velocities, as well as limited forces and collision detection.

\textit{P-Rob 3} is specifically designed to have easily interchangeable end-effectors to allow task-specific manipulation. Custom made mechanical interface requires only one screw to lock or release the connection. Integrated electrical pins pass through electrical signals as well as 24V power, which can then be adjusted using converters if needed. Additionally, a gigabit Ethernet cable is embedded in the robot arm. It allows establishing a connection between the devices in the gripper and the internal computers of Lio. 

\subsection{Gripper}
\label{subsec:sd_gripper}

Lio comes with a customised \textit{P-Grip} gripper, which has a friendly appearance and detachable magnetic eyes. The gripper has two custom-designed fingers with soft covers for grasping and lifting and manipulating objects up to $130$ mm.

One proximity sensor is embedded at the base of the gripper and each gripper finger has four infrared-based proximity sensors. They are used to detect whether an object is inside the gripper and measure the distance to the objects over and under the gripper fingers. It can be used for collision avoidance, surface and object detection as well as for interactive tasks like handshakes and entertainment. Covering the finger sensors by hand can also be used to confirm specific actions. Gripper fingers can be controlled using position and force controllers. Lio is also capable of learning objects using the sensor readings combined with finger positions to classify them later on during the grasping motion.

Gripper fingers are easily interchangeable. Four digital inputs and four digital outputs together with two $24$ VDC power outputs are available for the sensors.

Furthermore, there is an interchangeable camera module attached below the gripper fingers. It provides a video feed of $30$ frames per second (FPS).

For additional input options, an interactive ring can be mounted at the gripper, which has several programmable buttons. Also, \textit{Force Sensitive Resistor (FSR)} sensors are placed below the artificial leather of the gripper to detect physical presses on the head. 

\subsection{Software and Interfaces}
\label{subsec:sd_software}

\textit{P-Rob 3} is controlled by a \textit{C++} based low-level architecture called \textit{Motion Control Module (MCM)} and an optional \textit{Python}-based high level architecture \textit{myP}. The \textit{MCM} allows the immediate position and current control with a $100$ Hz update cycle via \textit{TCP} socket and includes all important safety features. There is also Python and \textit{C++} \textit{API} for the \textit{MCM}. \textit{myP} provides database, path planning and machine learning features which can be accessed through the robots user-friendly browser interface. The browser interface also includes a \textit{Python IDE} for applications development. Control of \textit{myP} is possible from any device using multiple interfaces such as \textit{TCP}, \textit{ROS} and \textit{Modbus (PLC)}. Vice versa, the robot can also control other machines via the same interfaces.

Different access levels can be set up on \textit{myP}. If a multi-access configuration is needed, each user can work according to the permissions granted.
Any scripts developed in \textit{Python IDE} can be copied between the robots and maintained using version control. The robot can be updated remotely for fixing bugs, updating functionality and adding new features.


\begin{figure}
    \vspace{0.2cm}
    \centering
    \includegraphics[width=0.5\textwidth]{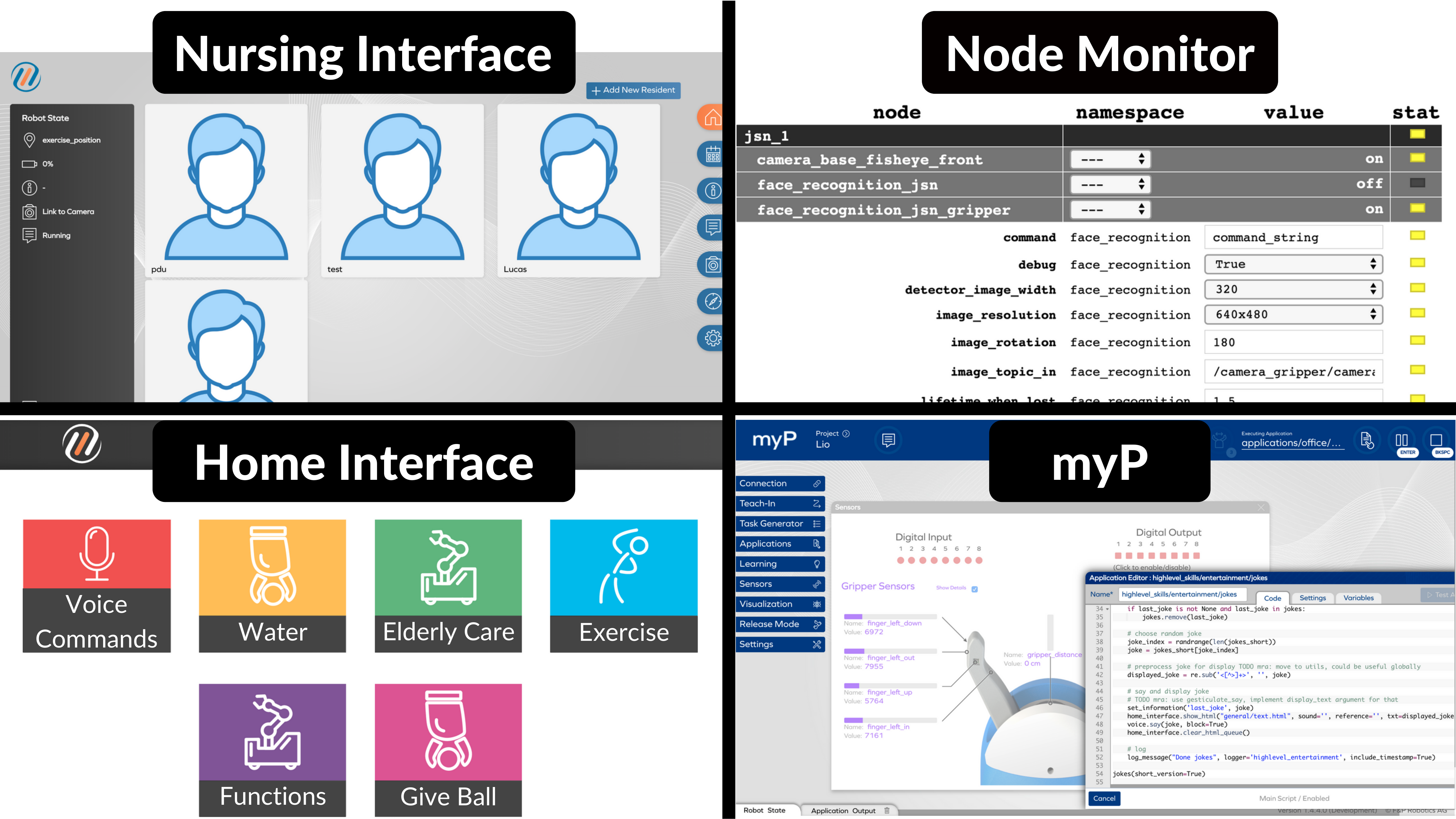}
    \caption{Available software interfaces for Lio: \textit{myP}, \textit{Nursing Interface}, \textit{Home Interface} and \textit{Node Monitor}. All of the interfaces are web-based and can be accessed from any browser-enabled device.}
    \label{fig:interfaces}
    \vspace{-0.6cm}
\end{figure}

The status of the robot can be monitored in real-time on multiple levels. It allows easy reconfiguration of Lio not only before the operation but also during runtime. For non-technical users, there are multiple simple interfaces with clearly visualised information. The \textit{Home Interface} provides high-level functions like launching specific scripts, switching autonomy mode on and off and remote control. For more detailed control, \textit{Nursing Interface} is used to schedule tasks and monitor system status in an easy to understand manner. It consists of the following information:
\begin{itemize}
  \item Location of the robot
  \item Battery level
  \item Current action and calendar
  \item History of actions and logs
  \item Real-time feeds of the cameras
  \item Patient data and teaching-in face recognition
  \item Remote control and manual steering of the robot
\end{itemize}

Additional sensor status information together with the script development and testing environment is available on \textit{myP}, specifically designed for advanced, technical users and developers. The connection between \textit{ROS} and \textit{myP} is established using messages and services providing full control of the robot from both interfaces. The \textit{Node Monitor} provides a summary of all the \textit{ROS} nodes that are running on the robot together with the ability to start, stop and restart each node individually and dynamically adjust parameters of the running processes. All of the interfaces can be accessed from any browser-enabled device. Additionally, all the standard \textit{ROS} tools and interfaces fully function on Lio for more in-depth debugging and development.
Examples of \textit{myP}, \textit{Nursing Interface}, \textit{Home Interface} and \textit{Node Monitor} are shown in Fig.~\ref{fig:interfaces}.

\subsection{Perception and Behaviour Algorithms}

Lio utilises sensors to enable behaviours used for assistance in everyday operations in health care institutions as well as for research purposes. Each of the algorithms can be enabled or disabled depending on the use case and overridden if needed. From the development perspective, it allows projects to reuse what is needed for the main operation of the robot while focusing on the development or improvement in specific areas.

Additionally, there are several AI algorithms deployed and fully integrated on Lio. These include:
\begin{itemize}
  \item Human Pose Estimation
  \item Object Detection and Recognition
  \item Object Grasping
  \item Face Detection and Recognition
  \item Door Opening and Closing
  \item Voice Recognition and Synthesis
\end{itemize}

Lio is equipped with a variety of sensors to get information about the environment. Developed algorithms make use of them by fusing necessary information to improve certain tasks such as grasping. To assist people grasping an object from the floor, the following information is used:
\begin{itemize}
    \item The object detection for classification of the object and its location in the camera image
    \item The 3D pointcloud of the RealSense camera for object localisation in the space
    \item The mobile platform's position for knowledge about the position of the platform in the environment
\end{itemize}
By fusing all this information, Lio is able to accurately drive to the target and move the robotic arm to grasp it. To improve the success rate of grasping, information of proximity sensors in the gripper fingers are used to fine-tune the gasping point and confirm successful grasp.

Currently, special markers need to be placed next to special objects for localisation in unstructured environments, for example, door handles, during the door opening task. However, the rest of the task, of observing if the door is open, trajectory calculation and opening action success is observed and confirmed by fusing visual and LiDAR data.

\subsection{Adaptations for COVID-19}

In the face of the \textit{COVID-19} pandemic, additional functionality was introduced to assist health care professionals in these extenuating circumstances. In contrast to other COVID-19 specific robots, Lio carries out this functionality next to its regular routines in health care institutions. Autonomous item delivery to staff and patients has already proven to be beneficial as it can be carried out in a contact-less manner.

Room disinfection is important measure to prevent the spread of the virus, especially in hospitals. Lio was adapted to carry out disinfection using an approved \textit{UV-C} light capable of effectively killing exposed germs, bacteria and viruses on the surfaces. The development focus was targeted at disinfection of frequently touched surfaces like door handles, light switches, elevator buttons and handrails. A custom holder for an \textit{UV-C} light was designed for Lio that can be carried on the back of the robot. At the current stage, \textit{ArUco} markers are placed next to the items to be disinfected. During the disinfection routine, Lio drives to dedicated locations on the map, locates the marker, grasps the \textit{UV-C} light and the places it over the object to be disinfected. During the operation, Lio indicates warning signs on the screen, LED light strip and gives a visual and verbal warning if a person is detected close to the robot. Exposure is limited only to the object of interest by the covered design of the holder.

One indicator of possibly infected people is Elevated Body Temperature (EBT). In the institutions where visitors are allowed, a common practice during the pandemic is to take body temperature measurements of people entering the building. To assist with that, remote measurement of passers-by was developed. The system is based on a thermal camera placed on the gripper, which is coupled with the colour gripper camera. Faces detected in the colour image are mapped to the thermal camera image. The point at the tear glands on the inside of the eye is located, as it has been proven to be a stable point with close to actual body temperature~\cite{ThermalI31:online}. For persons wearing non-transparent glasses, a second stable point, the maximum temperature in the area of the forehead is detected. EBT detection is done by comparing relative temperature differences between healthy people, who are recorded during the calibration process to the temperature detected of the passer-by. If EBT is detected with high confidence, a member of staff is notified to take a manual control measurement of the particular person with an approved medical thermometer.

\subsection{Error Handling}
To achieve a reliable performance of a complex system like Lio in an unpredictable environment, such as elderly care home requires a sophisticated error handling.

Error handling on Lio is divided into three main areas:
\textit{Low-level and hardware related} error handling is provided by \textit{myP} for the cases like arm collisions, activation of an emergency stop and sensor failure. For the arm collisions, multiple resolution options are available: pausing and resuming the movement or stopping the movement so that the skill fails followed by the next high-level skill resolving the issue.
\textit{Software components for sensing and perception} tasks are monitored and eventually restarted by operating system services and in the case of \textit{ROS nodes} by the \textit{Node Monitor}.
\textit{Higher-level skill-specific error} is handled in behavioural scripts. To explain the concept, an example use case of bringing a bottle of water to the person is taken. If a path is blocked while navigating to the person, the robot will actively ask a person for help. After the bottle is grasped from the inventory, a sanity check is performed by verifying the angle of gripper fingers in order to confirm the object was grasped. Upon failure, either the same motion is tried, or an alternative inventory slot is used to grasp the object again. If a person fails to take an object from Lio during the object handover, the robot would detect it and place the object back into the inventory.
\section{ROBOT OPERATION}
\label{sec:robot_operation}

Lio is an autonomous robot proactively performing tasks. On a high-level, this proactive autonomy is controlled by the decision engine system, which gathers information about the robot status and environment to choose the best-suited action to perform at a given moment~\cite{niemueller2013incremental}.

Operation of Lio is ensured by a combination of perception, memory and planning components.
Perception of the environment ranges from low-level sensory inputs up to high-level information delivered by AI algorithms. This real-time feed, together with introspective information and robot memory, serves as the information pool for the decision engine - the core component ensuring reactive autonomous behaviour of the robot. It is a high-level component deciding which behaviour should be executed in a particular situation.


The decision engine uses the \textit{SWI Prolog} logic programming language to model relations in the environment of Lio~\cite{wielemaker2012swi}. In addition to pre-defined common-sense rules, the users can define their reasoning mechanisms tailored to their application. All information available to the robot is evaluated by the set of decision rules to produce the most suitable action proposal at the given moment. This proposal is compared with inputs from manual user commands and scheduled actions from the calendar. Based on a priority system the most important action is selected for execution. This priority selection is also able to interrupt running skills to start more important ones. 


Commonly the deployment of Lio in a facility is connected with a set of project-specific routines. For example, in some institutions, Lio is distributing mail around the clinic and collecting blood samples to be delivered for analysis. A routine may require special equipment, material or tools. Lio can have custom holders for the tasks. The robot requests the staff to place a specific holder before starting the routine.

\subsection{Health Care Institution Requirements}
\label{subsec:ro_institution_needs}

Given the target robot deployment area being health care institutions, followed by home environments in the future, it is critical to comply with their requirements. The main concern when discussing robot deployment in health care institutions is data protection. Hospitals, rehabilitation clinics or elderly homes do have concerns about having sensitive information in the institution. With several cameras on the robot, it has to be ensured that any unwanted observations do not take place. It includes remote monitoring of the sensor data, storing or sending out any visual information.

To comply with the privacy requirements, Lio was designed to have all the visual and navigation data processed by on-board computers. If any information needs to be stored, for example, learning face recognition data, it is anonymised and encoded in a way that original images cannot be recovered. At the moment, only voice data is processed off-board, but a data protection contract is signed with companies providing these services. Furthermore, it is possible to set up a safe \textit{Virtual Private Network (VPN)} connection to access and support Lio remotely.

For full-feature operation, Lio needs a secure WiFi connection, which typically is present in many health care institutions. Lio can connect to existing networks, so the IT department of the institution sets security measures. The robot can operate without a wireless connection, but access to the interfaces will not be possible.

Another common request is to prevent the robot from driving to restricted areas. Allowed and no-go areas can be defined after creating a floor map of the institution and adjusted as needed later on. Also, it is possible to define these areas for different times of the day depending on the task Lio is executing.

Integration to existing hospital systems, like phones or nursing call system is also requested. However, the existing standards depend on the systems used, and it can be challenging to have a universal solution. Given that the system used has an existing \textit{API}, integration over \textit{TCP} or scripting language, it can be done individually for each institution.

\subsection{Lio Deployment}
\label{subsec:lio_deployment}

Deployment of Lio in health care institutions consists of several steps, including preparation. First of all, the facility is analysed by indicating the areas of operation for the robot, evaluating the navigation and traversability capabilities and recording the areas, usually by taking photos of the floor, hallways, doors, rooms and handles that the robot will need to operate. Then, the specific tasks of the robot are discussed and decided upon with the client. The developer will adapt the existing functionalities and develop new functions. The client prepares the WiFi network for Lio according to their privacy requirements and making sure the robot can connect and has good coverage along the operational routes. Furthermore, a responsible person in the facility for operating Lio is assigned.

Once Lio is delivered, the developer will install the robot. One of the first steps is to create a map of the facility and to find a designated charging spot for the robot. The teaching of the staff is done in two steps. One is to introduce staff members of the facility to the robot, explain the capabilities and limitations of the robot. Then, a selected number of employees are given more extensive training on how to operate the robot, restart, manually charge it if needed and a full introduction to \textit{Nursing} and \textit{Home Interfaces} for controlling the robot. It is common to have a small \textit{Welcome Lio} event involving residents and patients in the facility. From then on, Lio starts working autonomously in the facility.

Lio deployment in research institutions follows similar steps. However, depending on the project goals of the lab acquiring the robot, the robot is adapted to the needs and training plan is tailored according to the development and use plans. For the labs focusing on in-depth algorithm development, a thorough introduction to \textit{myP}, scripting and ROS integration is provided.

\section{EXPERIMENTS AND USE CASES}
\label{sec:experiments}

Lio robots have been deployed in seven different health care institutions in Germany and Switzerland. Some robots already operate for over a year. Robots have daily routines such as collecting lab samples from different wards and bringing them to the pick-up point at the reception and then delivering the mail to those same wards in a rehabilitation clinic. In the remaining time, Lios entertain patients and visitors with stories of former patients or encourages to join in on a few simple exercises. Robots also remind residents about their scheduled activities by knocking on their door, opening it, saying out loud information about upcoming events and closing the door when leaving. Staff can control Lio using a tablet to schedule reminders and start high-level commands like playing the music and offer snacks. Robots are also capable of driving to the rooms of the residents and taking the menu orders. This information is sent to the catering service.

\begin{figure}[ht]
    \centering
    \includegraphics[width=0.49\textwidth]{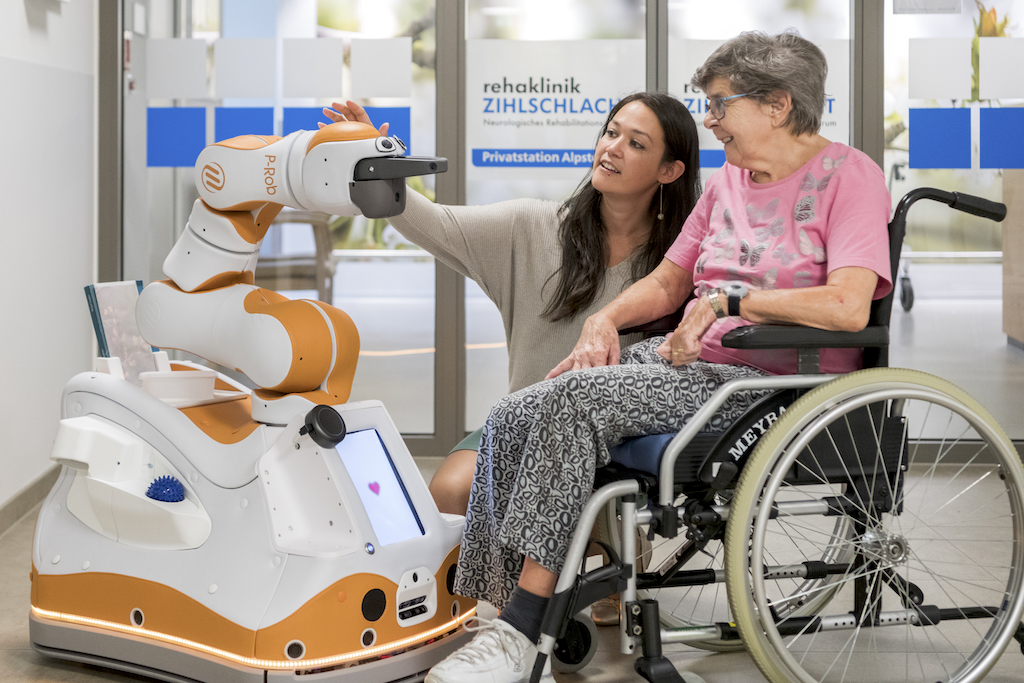}
    \caption{Patients and staff are interacting with Lio in a rehabilitation clinic in Switzerland.}
    \label{fig:lio_clinic}
    \vspace{-0.2cm}
\end{figure}

In one of the projects, Lio was deployed in the home of a person with paraplegia for several weeks. The robot allowed the woman to have more independence by assisting with her daily tasks like taking and opening a bottle using a built-in automatic bottle opener and handing it over. She could manually control Lio over her smartphone to assist with more complex tasks such as taking off the jacket.


\subsection{Usability Evaluation}
\label{subsec:usability_evaluation}


According to the guidance document for medical devices by the FDA, development of the behaviours of Lio was guided by human factors and usability engineering~\cite{food2016applying}. Contextual inquiries, interviews and formative evaluations were conducted to evaluate all the existing functions~\cite{wirth2020erfahrungen}~\cite{fruh2018erfahrungen}~\cite{gasser2017bodily}. Following the simulated testing at the lab, usability tests were conducted with the actual end-users in their natural environment.

In a user study on co-robots, it was determined that the introduction of an innovative device such as a robot in health care leads to unexpected challenges~\cite{bendel2020co}.
One finding was that the head which is sometimes used to interact with Lio was not always reachable for a person in a wheelchair due to the distance. For safety reasons, the head of Lio was programmed not to extend beyond the base of the mobile platform which impeded interaction with the robot~\cite{wirth2020erfahrungen}. 

Operational data was collected in one of the rehabilitation clinics where Lio was deployed. On a monthly basis, for delivery tasks, Lio drove on average $16.8$ km for delivery tasks and additionally $25.2$ km during entertainment tasks. Entertainment functions were triggered 96 times per month on average. The success rate of the autonomous daily delivery function was $85.5\%$ in the period from February 1 to May 8, 2020. In total there were $186$ planned deliveries of transporting mail and blood-samples around the clinic. The most common reasons for failure were device failure of low-level sensors, charging issues, and network connection problems. There were also errors such as the pressed emergency stop not being detected.

Generally, elderly people and health care staff are very curious and open towards the robot. People are polite to Lio even though they are aware that it is a machine~\cite{gasser2017bodily}. Also, multiple usability tests and qualitative interview studies have been conducted to improve the interaction patterns. Supporting findings from other home trials~\cite{fischinger2013hobbit}~\cite{broadbent2009acceptance}~\cite{syrdal2008sharing}, patients expected the robot to have a certain personality. During an ethical evaluation of the robot's deployment, an elderly person described Lio as a play companion. He consciously engaged in a game with Lio multiple times, giving an impression that it is a sentient being. The robot gets often patted on the head by passersby.

On the interaction side, additional studies were done on the topic of robot speech. Despite female voices being perceived more friendly according to the studies, Lio is programmed to speak using a male voice~\cite{broadbent2012attitudes}. A deeper male voice was easier to understand for elderly people. If the robot speaks in a realistic human voice, users are more likely to reply to the robot using voice over other types of inputs~\cite{gasser2017bodily}. Typically the pronunciation and conversations used by Lio are simplified, which in turn resulted in a higher response rate.

\subsection{Current Limitations}

New challenges are constantly observed during the deployments of Lio in real-life situations. Even though combined voice and tactile inputs have proven to be a suitable option, in certain situations, it is difficult for Lio to reliably understand some people with disabilities as well as some elderly people. Also, navigation in cluttered environments and some narrower places not adapted for wheelchairs can be challenging. 

Furthermore, depth camera is not currently present in the gripper, making precise picking of the objects placed higher up, like a tabletop more complicated compared to the picking from the floor. For some specific manipulation actions, like a door opening, markers are still needed. Language understanding is currently limited to keyword matching. Language understanding service still relies on online service, thus constant listening is not used due to privacy concern.

Currently, Lio does not proactively approach people for interaction. Because of this, a reduction of interest in the robot was observed among some people as time passes. At the moment, additional functionality is gradually introduced as it is developed during the lifetime of the robot to keep people interacting with Lio and discovering new behaviours.
\section{CONCLUSIONS AND FUTURE WORK}
\label{sec:conclusions}

Lio is considered an all-in-one platform suitable for human-robot interaction and personal care tasks. Its design has evolved, both in hardware and software, to address the limitation of similar platforms and ensure the requirements of health care institutions, as well as home care, are met. Lio is intrinsically safe due to the padded artificial-leather covers covering the majority of the robot, limited forces and speed, soft mode, advanced navigation and behaviour algorithms. The robot complies with \textit{ISO13482} standard - safety requirements for personal care robots. Lio has a robotic arm placed on a moving platform; it is capable of both, assisting and interacting with people.

Lio has been proven during multiple deployments in health care institutions and had positive feedback regarding functionality as well as acceptance by patients, residents and medical staff. Currently, the majority of the robots were sold as products, while the remaining deployments are part of the on-going projects. Autonomous operation capability ensures that Lio can be easily deployed with just a brief training time for staff on how the robot operates and how it should be used.

What differentiates Lio from other service robots are the manipulation capabilities of the large arm and gripper which allows opening doors, grasping, bringing, and handing over water bottles, and handling tools for UV-C disinfection. Combined with large and customisable inventory space, it allows Lio to carry and operate a large variety of tools.

All the basic functionalities, as well as some advanced navigation, perception and AI algorithms, are deployed on the robot. Algorithms can be easily enabled and disabled depending on the project needs making it easy for researchers and developers to use existing algorithms and focus on their field of expertise for improvements.

Having multiple user interfaces makes Lio usable by both, tech-savvy and inexperienced users and allows them to easily observe the status of the robot, schedule and adjust the behaviour according to the needs. With a possibility of having a TCP communication, integration can be implemented with external systems.

According to customer feedback, Lio is constantly being updated both, in terms of hardware and software. The development is focused to improve upon identified issues and to increase the level of autonomy. Multi-floor mapping and elevator use is planned, as well as marker-less identification and localisation of objects like door handles, elevator buttons and light switches to allow the robot operation in larger areas of the facilities. Lio can manipulate a variety of objects and have them in inventory. Motion planner will be adapted to take items held on the gripper or on the back of the robot into consideration when calculating the collision-free path. Another goal is to make Lio more proactive in terms of interaction. It means advanced behaviour models to allow the robot to find specific people around the facility, actively approaching them and enabling scene understanding to determine if certain expected actions occurred. For example, if a glass of water was handed to a person, to ensure that he or she drank the water to stay hydrated.

Additionally, hardware adjustments are planned to include advanced 3D sensor in the gripper, as well as to update the appearance to convey emotional responses in user communication. Improvements on natural language processing are planned as well to enhance the communication with wake-word recognition and chatbot capabilities.

To enhance the robot behaviour development, improve testing and evaluate capabilities of Lio before acquiring the robot, a full-feature realistic simulation is being developed with full support of \textit{ROS} and \textit{ROS2}, including all the interfaces described in this paper.

\section*{ACKNOWLEDGMENT}

Authors would like to acknowledge the hard work and dedication of all \textit{F\&P Robotics AG} employees and partners in developing and testing a personal assistant robot Lio.






\bibliographystyle{IEEEtran}
\bibliography{IEEEexample}

\end{document}